\documentclass[sigconf,screen,nonacm]{acmart}
\AtBeginDocument{%
  }
\usepackage{amsfonts}
\usepackage{cuted,lipsum}
\usepackage[most]{tcolorbox}
\usepackage{subcaption}
\usepackage{tabularx}
\usepackage{graphicx}
\usepackage{hyperref}
\usepackage{multirow}
\usepackage{array}
\usepackage{balance}
\usepackage{makecell}
\usepackage{soul}
\setcopyright{acmlicensed}
\copyrightyear{2025}
\acmYear{2025}
\acmDOI{XXXXXXX.XXXXXXX}
\acmConference[MM'25]{}{October 27--31,
  2018}{Dublin, Ireland}
\acmISBN{978-1-4503-XXXX-X/2018/06}

\begin{document}

\title{Multimedia Verification Through Multi-Agent Deep Research Multimodal Large Language Models}

\author{Huy Hoan Le}
\authornote{Both authors contributed equally to this research.}
\email{hoanlh4@fpt.com}
\affiliation{%
  \institution{Quy Nhon AI, FPT Software}
  \city{Quy Nhon}
  \country{Vietnam}
}

\author{Van Sy Thinh Nguyen}
\authornotemark[1]
\email{thinhnvs@fpt.com}
\affiliation{%
  \institution{Quy Nhon AI, FPT Software}
  \city{Quy Nhon}
  \country{Vietnam}
}

\author{Thi Le Chi Dang}
\email{chidtl@fpt.com}
\affiliation{%
  \institution{Quy Nhon AI, FPT Software}
  \city{Quy Nhon}
  \country{Vietnam}
}

\author{Vo Thanh Khang Nguyen}
\email{khangnvt1@fpt.com}
\affiliation{%
  \institution{Quy Nhon AI, FPT Software}
  \city{Quy Nhon}
  \country{Vietnam}
}

\author{Truong Thanh Hung Nguyen}
\email{hung.ntt@unb.ca}
\authornote{Corresponding author.}
\orcid{0000-0002-6750-9536}
\affiliation{%
  \institution{University of New Brunswick}
  \city{Fredericton}
  \state{New Brunswick}
  \country{Canada}}

\author{Hung Cao}
\email{hcao3@unb.ca}
\orcid{0000-0002-0788-4377}
\affiliation{%
  \institution{University of New Brunswick}
  \city{Fredericton}
  \state{New Brunswick}
  \country{Canada}}

\begin{abstract}
    This paper presents our submission to the \textit{ACMMM25 - Grand Challenge on Multimedia Verification}. We developed a multi-agent verification system that combines Multimodal Large Language Models (MLLMs) with specialized verification tools to detect multimedia misinformation. Our system operates through six stages: raw data processing, planning, information extraction, deep research, evidence collection, and report generation. The core Deep Researcher Agent employs four tools: reverse image search, metadata analysis, fact-checking databases, and verified news processing that extracts spatial, temporal, attribution, and motivational context. We demonstrate our approach on a challenge dataset sample involving complex multimedia content. Our system successfully verified content authenticity, extracted precise geolocation and timing information, and traced source attribution across multiple platforms, effectively addressing real-world multimedia verification scenarios.
\end{abstract}





\maketitle

\section{Introduction}
Multimedia misinformation has become a critical challenge in the digital information landscape, with visual content serving as a primary vector for spreading false narratives. Recent studies show that around 80\% of fact-checked misinformation cases involve images or videos, making multimedia verification essential for maintaining information integrity \cite{duwal2025evidence}. Modern misinformation employs two main strategies: deepfakes involving sophisticated content manipulation, and cheapfakes where genuine media is repurposed in misleading contexts.
Current multimedia verification approaches typically focus on either technical forensics or contextual analysis in isolation. Traditional media forensics tools excel at detecting pixel-level manipulations but struggle with context verification, while text-based fact-checking systems cannot adequately process visual information. Although Multimodal Large Language Models (MLLMs) offer new possibilities for multimedia analysis, they face challenges with hallucination and lack grounding in verifiable evidence sources.

The \textit{ACMMM25 - Grand Challenge on Multimedia Verification} \cite{acmmm25-grand} provides an opportunity to address these limitations through comprehensive verification systems that can handle real-world misinformation scenarios. The challenge requires systems to process complex multimedia content and provide detailed verification reports with evidence-based assessments of authenticity and contextual accuracy. In this paper, we present a multi-agent multimedia verification system designed specifically for this challenge. Our approach integrates MLLMs with specialized verification tools through a systematic six-stage pipeline that processes multimedia content from initial data extraction through comprehensive report generation. Our contributions are as follows:
\begin{itemize}
    \item \textbf{Multi-agent verification architecture:} We propose a six-stage pipeline that systematically processes multimedia content from data extraction through comprehensive report generation, ensuring thorough coverage of both technical and contextual verification requirements.
    \item \textbf{Deep Researcher Agent with specialized tools:} We introduce an agent that employs four verification tools, including reverse image search, metadata analysis, fact-checking databases, and a novel verified news processor that extracts spatial, temporal, attribution, and motivational context.
    \item We demonstrate the effectiveness of our proposed system through evaluation on the \textit{ACMMM25 - Grand Challenge on Multimedia Verification} dataset \cite{acmmm25-grand}, successfully handling complex multimedia content involving geopolitical events.
\end{itemize}

\begin{figure}
    \centering
    \includegraphics[width=.9\linewidth]{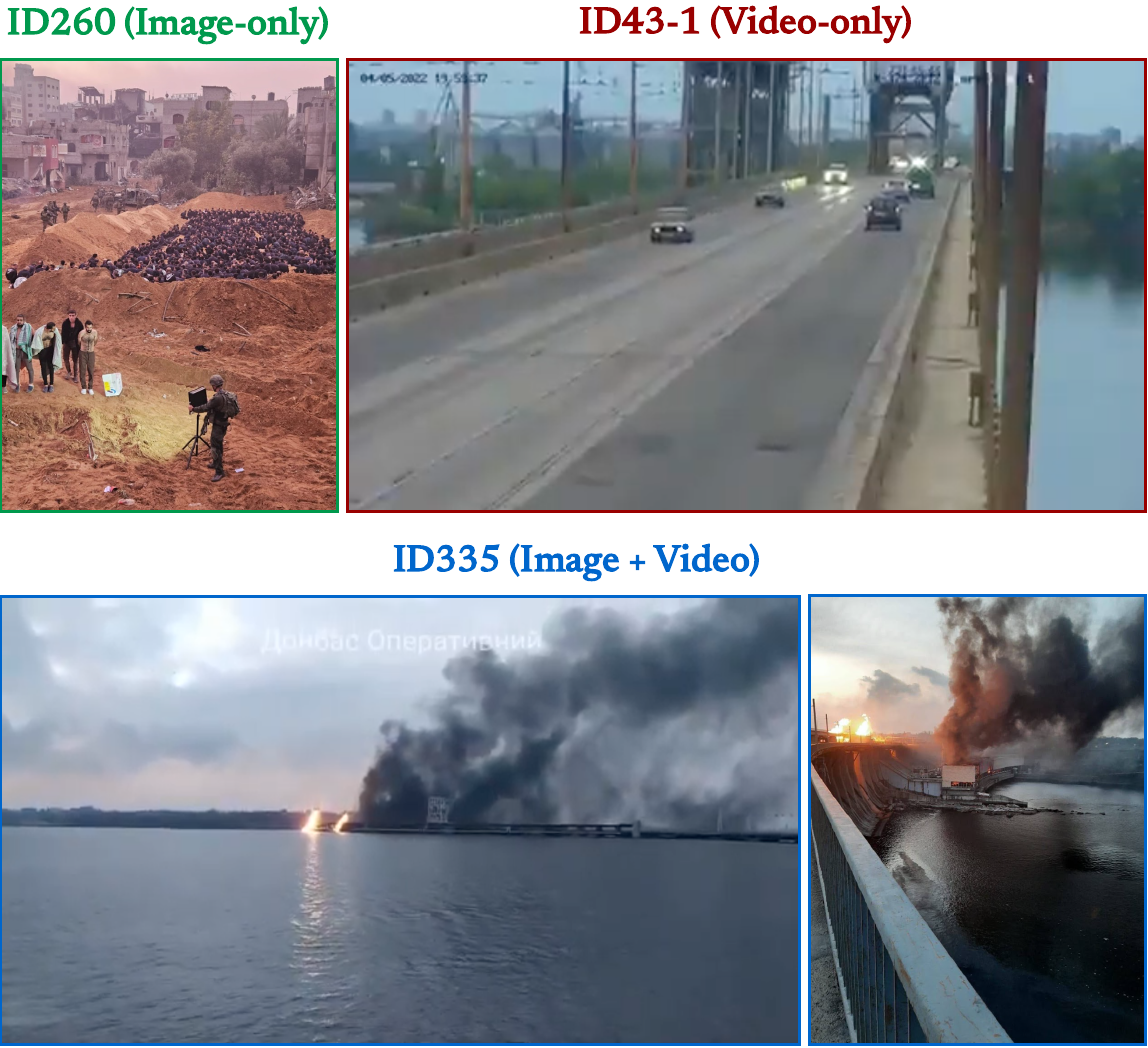}
    \caption{\textit{ACMMM25 - Grand Challenge on Multimedia Verification} dataset for verifying the authenticity and context of multimedia content.}
    \label{fig:sample}
\end{figure}
\section{Related Work}
\subsection{Multimedia Verification}
The field of multimedia verification has experienced substantial growth in recent years, driven by the increasing prevalence of misinformation and the sophisticated nature of modern content manipulation techniques. As visual content becomes a major vector for misinformation, research on verifying the authenticity and context of images/videos has grown significantly. Recent studies found that around 80\% of fact-checked misinformation cases online include an image or video \cite{duwal2025evidence}. 

Prior studies highlight two major challenges: (1) ``deepfakes'' (i.e., content tampering and synthesis, where images/videos are digitally altered or generated) and (2) ``cheapfakes'' (i.e., content miscontextualization, where genuine media is reused in a false context to mislead). Early efforts in media forensics tackled image manipulations like splicing, copy-move, or GAN-generated fakes. For example, the MediaEval 2016 Verifying Multimedia Use task aimed to automatically detect manipulated and misleading use of web images and videos \cite{boididou2015verifying}. They defined a post as fake if the visual content does not actually depict the event described by the accompanying text, e.g., an old or unrelated photo falsely presented as current news. Large benchmark datasets such as the NIST Media Forensics Challenge \cite{guan2025nist} and FaceForensics++ \cite{rossler2019faceforensics} have driven progress in detecting tampered images and AI-generated videos. In parallel, the recent ``Detecting Cheapfakes'' Grand Challenges at ACM Multimedia and related venues specifically target out-of-context (OOC) image misuse, reflecting a growing research emphasis on context verification alongside traditional media forensics \cite{aneja2022acm}.

% \subsubsection{Evidence-Based Verification} 

\subsection{Multimodal LLMs (MLLMs) for Multimedia Verification}
Given the inherently multimodal nature of online misinformation, recent research has trended toward joint image–text verification.
Prior studies have shown that leveraging both the visual and textual modalities can improve accuracy over text-only or image-only methods \cite{tran2022textual,nguyen2023leveraging,nguyen2023multi,braun2024defame,gao2024knowledge,nguyen2024hybrid,liu2024fka,heidari2024deepfake,kumar2025advances}. Previous attempts at multimedia verification have focused on developing comprehensive Open Source Intelligence (OSINT) methodologies combining reverse image search platforms \cite{sonal2016,ganti2022novel,alqudah2023osint,meredith2024osint}, metadata analysis tools \cite{ganti2022novel}, structured fact-checking frameworks \cite{nguyen2024hybrid}, and knowledge graphs \cite{dao2023leveraging,nguyen2025multimodal,duwal2025evidence}.

The advent of MLLMs has opened new avenues for multimedia verification. 
These models bring vast parametric world knowledge from their training data, which can help recognize when a caption makes implausible claims about an image.
While MLLMs excel at flexible reasoning, they have drawbacks. A known issue is hallucination, where a MLLM may fabricate plausible-sounding details if it lacks actual evidence \cite{ghosh2024logical,kim2024can,xie-etal-2024-v}. This is problematic in fact-checking, as the system might incorrectly justify a decision with false information.

To counter this, recent research has explored hybrid approaches that combine LLM reasoning with external knowledge retrieval or structured representations, such as treating the verification task as a multi-step reasoning process orchestrated by AI agents. \citet{lakara2024mad} involved an agent or multiple agents that can dynamically query tools and data sources, e.g., performing image analysis, web searches, and knowledge base lookups, guided by an LLM’s logic. \citet{duwal2025evidence} integrated information from knowledge graphs or used graph neural networks to ensure decisions are grounded in real data. Such methods aim to retain the interpretability of LLM-based reasoning while improving accuracy and trustworthiness by anchoring the model’s output to verifiable evidence. Also, \citet{liu2024fka} developed FKA-Owl, a framework that leverages forgery-specific knowledge to augment MLLMs, enabling them to reason about manipulations effectively by incorporating semantic correlations between text and images and artifact traces in image manipulation. Similarly, \citet{gao2024knowledge} proposed a knowledge-enhanced vision and language model that integrates information from large-scale open knowledge graphs to augment the ability to discern the veracity of news content. Such multimodal LLM agents can dynamically choose operations (e.g., reverse image search, object recognition) and reason about the results in a conversational manner. \citet{braun2024defame} introduced DEFAME, a zero-shot multimodal fact-checking pipeline that uses an LLM backbone to retrieve and analyze text and image evidence, producing structured verification reports.
\citet{nguyen2024hybrid} introduced a two-stage cheapfake-detection pipeline that first performs fast reputation checking by reverse-image-searching a photo and verifying whether it appears on trusted news domains, then passes the image–caption triplet to a visual-language network trained primarily on LLM-generated synthetic data to judge contextual integrity.

Our work follows this trend, focusing on the main task of multimedia verification by leveraging an MLLM agent that can analyze content and use online information. Our agent-based approach builds upon prior research by marrying the strengths of MLLM reasoning and tool-assisted verification, aiming for high accuracy and reliability in tackling real-world misinformation challenges.

\begin{figure*}[th!]
    \centering
    \includegraphics[width=\linewidth]{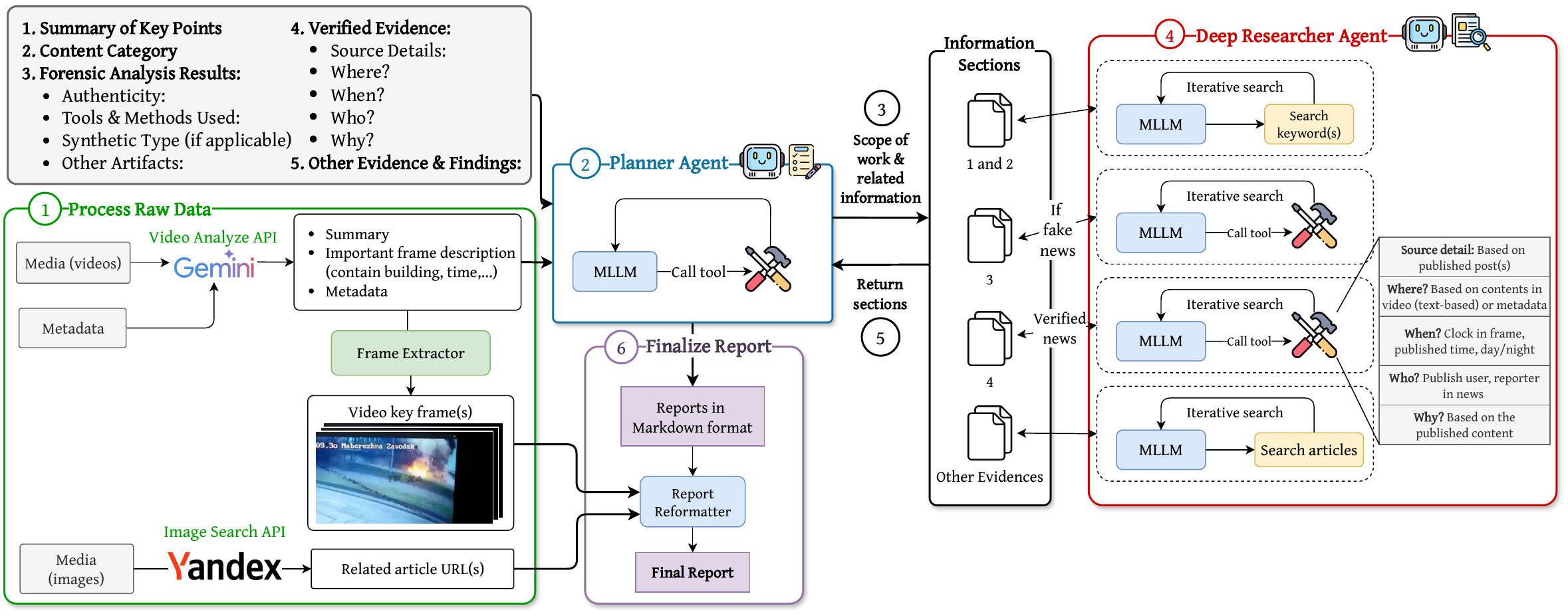}
    \caption{Our proposed Multi-Agent Deep Research MLLMs architecture for the multimedia verification system.}
    \label{fig:arch}
\end{figure*}

\section{Dataset}
In this paper, we tackle the main task of the \textit{ACMMM25 - Grand Challenge on Multimedia Verification} \cite{acmmm25-grand}. The dataset comprises a collection of 50 high-quality samples designed to reflect real-world multimedia verification challenges. Each sample contains multimedia content files including videos in \texttt{.mp4} format and images in \texttt{.jpg} format, with audio and video data constituting 72\% of the total dataset, while static images account for the remaining. 
While in the validation set, as shown in Figure~\ref{fig:sample}, most data samples are video-only, case \texttt{ID260} is image-only, and case \texttt{ID335} contains both video and images.

Each verification case is provided as a comprehensive package including: (1) Primary multimedia content consisting of image(s) or video(s) that may originate from various sources and potentially feature content in languages other than English; (2) Contextual information such as captions, descriptions, social media posts, news articles, and available metadata that provide background for the multimedia content; and (3) Additional investigative clues including possible source attributions, claims made about the content, or preliminary fact-checker notes when relevant to the case.

The dataset requires participants to produce comprehensive verification that systematically addresses multiple dimensions of content authenticity and context. The verification framework encompasses three primary analytical components: temporal and spatial verification requiring precise determination of location (geological coordinates) and timing of events (specific dates); forensic authenticity analysis involving detection of synthetic content, modifications, or recapturing using specialized tools and techniques; and evidential cross-verification demanding corroboration through both the provided multimedia data and external internet sources.

Importantly, the challenge's objective extends beyond binary authenticity determination. Instead, it emphasizes comprehensive evidence assessment that identifies which specific elements of the multimedia content can be independently verified through external sources and which aspects remain unverifiable or inconclusive. This approach recognizes the nuanced nature of multimedia verification in real-world scenarios where complete verification may not always be feasible, requiring participants to clearly distinguish between confirmed facts, uncertain elements, and information that cannot be substantiated through available evidence.
\section{Methodology}

Our multimedia verification system employs a multi-agent architecture that combines MLLMs with external verification tools to systematically analyze and fact-check multimedia content. The system operates through six distinct stages: (1) raw data processing, (2) planning and coordination, (3) information extraction and sectioning, (4) deep research and verification, (5) evidence collection, and (6) comprehensive report generation, as outlined in Figure~\ref{fig:arch}.

\subsection{Stage 1 – Raw Data Processing}
The initial stage handles diverse multimedia inputs through specialized processing pipelines designed for different media types.

\subsubsection{Video Processing Pipeline} For video content, we utilize the Gemini 2.0 Flash as the core MLLM to analyze videos and extract comprehensive metadata and contextual information. The system generates frame-by-frame descriptions that capture temporal dynamics, identify key objects and scenes, and extract technical metadata including timestamps, resolution, and encoding information. A Frame Extractor component automatically identifies and extracts the most informative keyframes that represent critical visual moments in the video sequence.

\subsubsection{Image Processing Pipeline} Static images are processed through the Yandex Image Search API \cite{cloud_2025}, which performs image searches to identify potential source materials, related articles, and previous uses of the image across the web. This component generates a comprehensive list of related article URLs and metadata that forms the foundation for subsequent verification steps.

\subsection{Stage 2 – Planner Agent}
The Planner Agent serves as the central coordination hub, implemented using an LLM with tool-calling capabilities. This agent analyzes the processed multimedia data and associated metadata to develop a systematic verification strategy. The planner determines which verification tools and methods are most appropriate for the specific content type and potential misinformation vectors identified during initial processing.
The Planner Agent organizes the verification workflow by creating structured information packets that guide subsequent analysis stages. It identifies key claims, potential inconsistencies, and areas requiring detailed investigation, then delegates specific verification tasks to specialized components.

\subsection{Stage 3 – Information Extraction and Sectioning}
Based on the planner's strategic assessment, the system extracts relevant information and organizes it into discrete, analyzable sections. Each section contains specific aspects of the multimedia content that require independent verification, such as:
\begin{itemize}
    \item \textbf{Temporal Claims:} Date and time assertions made by or about the content.
    \item \textbf{Geographical Claims:} Location-based information and spatial context.
    \item \textbf{Entity Recognition:} People, organizations, or objects featured in the content.
    \item \textbf{Contextual Metadata:} Technical and social context surrounding content creation and distribution.
\end{itemize}
This sectioning approach enables parallel processing of different verification aspects while maintaining systematic coverage of all potentially misleading elements.

\subsection{Stage 4 – Deep Researcher Agent}
The Deep Researcher Agent represents the core verification engine, implementing an iterative search and analysis framework. This agent employs multiple search strategies:
\begin{itemize}
    \item \textbf{Keyword-Based Search:} The agent generates contextually relevant search terms based on content analysis and performs systematic web searches to gather corroborating or contradicting evidence.
    \item \textbf{Tool-Assisted Verification:} The system integrates multiple external verification tools, including reverse image search engines, metadata analysis utilities, and fact-checking databases. When processing verified news content, this tool systematically extracts four critical source details:
    \begin{itemize}
        \item Source detail (details of the published posts),
        \item Where? (spatial context based on contents in video text, or metadata), 
        \item When? (temporal context using clocks in frames, published time, day/night indicators, 
        \item Who? (attribution context identifying published users and reporters in news), \item Why? (motivational context based on the published content analysis).
    \end{itemize}\item \textbf{Source Analysis:} For each piece of evidence discovered, the agent performs detailed source verification, examining publication details, publisher credibility, temporal consistency, and cross-referencing with multiple independent articles.
\end{itemize}
The Deep Researcher Agent maintains detailed provenance tracking, documenting the complete chain of evidence discovery and analysis decisions to ensure transparency and reproducibility.

\subsection{Stage 5 – Evidence Collection and Synthesis}
The system aggregates findings from all verification components, organizing evidence according to reliability, relevance, and consistency. This stage performs aggregation of different evidence sources and identifies potential conflicts or contradictions that require additional investigation.
Evidence is categorized into verified facts, related information, and disputed claims, with confidence scores assigned based on source reliability and evidence consistency. The system also identifies gaps in available evidence and areas where verification remains inconclusive.

\subsection{Stage 6 – Report Generation and Formatting}
The final stage synthesizes all verification findings into a comprehensive, structured report. The Report Generator creates detailed documentation following a standardized format:
\begin{itemize}
    \item \textbf{Executive Summary:} Key findings and overall assessment of content authenticity and contextual accuracy.
    \item \textbf{Content Classification:} Categorization of the multimedia content type and potential misinformation vectors identified.
    \item \textbf{Forensic Analysis Results:} Technical analysis of content authenticity, including manipulation detection results, metadata analysis, and synthetic content identification where applicable.
    \item \textbf{Verified Evidence Documentation:} Comprehensive source attribution including temporal, geographical, and contextual details with complete provenance tracking.
    \item \textbf{Additional Findings:} Supplementary evidence, related content analysis, and potential implications for broader misinformation patterns.
\end{itemize}
The Report Reformatter ensures consistent formatting and presentation, producing both human-readable summaries and machine-readable structured outputs suitable for integration with broader fact-checking workflows.

Altogether, our system leverages MLLMs for reasoning and coordination, while maintaining grounding through external tool integration and evidence-based verification. The modular architecture enables scalable processing of diverse multimedia content while maintaining systematic verification standards. The iterative nature of the Deep Researcher Agent allows for adaptive investigation strategies that can handle novel misinformation patterns and evolving content manipulation techniques.

\section{Demonstrative Results}
\newtcolorbox[auto counter]{prompttemplate}[1][]{
  enhanced,
  fonttitle=\scshape,
  #1
}
\definecolor{ForestGreen}{RGB}{34,139,34}
\begin{figure*}[h!]
\begin{prompttemplate}[label=template:A,
  title={\texttt{ID43-3}: Verificiation Report}
]
\footnotesize
\textbf{\# Case Summary:}
Missile Strike on Dnipro Bridge Verification Report
\begin{itemize}
    \item Overview: Multiple video sources and associated metadata depict a missile strike on a bridge in Dnipro, Ukraine. The footage, reportedly recorded from a hotel, shows the explosion at or near Naberezhna Zavodska, with visible aftermath including a significant fireball, rising smoke, and infrastructural damage to a bridge frequently used by vehicles (and possibly trams). Data overlays in the video provide location details and timestamps. 
    \item Verification Status: Verified
    \item Geolocation Coordinates: Approx. 48.4647° N, 35.0462° E (Dnipro, Ukraine – approximate central coordinates)
    \item Exact Date: 04/05/2022, with one video timestamp noted at 19:58:37 (local time)
\end{itemize}

\textbf{\# Content Classification}
\begin{itemize}
    \item \textbf{Tags:} Content Classification:
    \item Platforms: Twitter (social media video post), Multiple Video Sharing Platforms
    \item Involved Entities: Hotel occupants filming, local eyewitnesses, potential military actors (inferred from combat context)
    \item Event Type: Missile strike in a combat zone
    \item Content Category: Combat actions and high violence explosive event
\end{itemize}

\textbf{\# Verified Evidence}
\begin{itemize}
\item \textbf{Source Details:}
    \begin{itemize}
        \item Original Publisher: The footage was initially shared via a Twitter account\\ (\texttt{https://twitter.com/cesarnews4/status/1521949838916521984}) and subsequently reshared on various video platforms.
        \item Platform Specifics: The videos contain embedded overlays showing details such as time (e.g., 9:30 and a timestamp reading '04/05/2022 19:58:37') and location (Naberezhna Zavodska, Dnipro) that corroborate the metadata.
    \end{itemize}
\item \textbf{Where? (Location):}
    \begin{itemize}
        \item Verified Location: Naberezhna Zavodska in Dnipro, Ukraine. Visual confirmation through recognizable landmarks (bridge structure and urban skyline) aligns with known images of Dnipro.
       \item Approximate Coordinates: 48.4647° N, 35.0462° E.
    \end{itemize}  
\item \textbf{When? (Time):}
    \begin{itemize}
        \item Verified Timestamps: Overlays and metadata in the videos indicate the event occurred on 04/05/2022 with one video showing an exact timestamp of 19:58:37 local time. Additional visual cues (shadow positions, morning/evening light differences) support temporal claims.
    \end{itemize}
\item \textbf{Who? (Entities Involved):}
    \begin{itemize}
        \item Individuals: The videos appear to be captured by hotel occupants or local eyewitnesses.
        \item Groups: Indirect evidence suggests involvement of military forces executing a missile strike.
    \end{itemize}
\item \textbf{Why? (Motivation or Intent):}
    \begin{itemize}
        \item Evidence-Based Assessment: The content was created to document a military/missile strike and highlight the ensuing damage in a combat zone, likely aimed at informing the public and possibly influencing opinion regarding the conflict in Ukraine.
    \end{itemize}
\end{itemize}

$\texttt{\![Evidence Image](report/ID43-2.mp4\_00\_02\_2.50s.jpg)}$

$\texttt{\![Evidence Image](report/ID43-2.mp4\_00\_02\_to\_00\_11\_3.50s.jpg)}$

$\texttt{\![Evidence Image](report/ID43-1.mp4\_00\_00\_to\_00\_02\_2.00s.jpg)}$

\textbf{\# Forensic Analysis}
\begin{itemize}
    \item \textbf{Authenticity Assessment:} Detailed frame analysis confirms the presence of embedded overlays (time, location, MAC-address information) that are consistent across multiple video segments, indicating the footage is not synthetically manipulated. The consistency in lighting, shadow patterns, and environmental cues further supports the video's authenticity.
    \item \textbf{Verification Tools \& Methods:}
        \begin{itemize}
            \item Frame-by-frame analysis and timestamp cross-referencing using digital forensic software.
            \item Metadata inspection of video files confirming the overlay data (timestamps, location detail).
            \item Comparison with satellite imagery and known landmarks of Dnipro to verify visual landmarks.
        \end{itemize}
    \item \textbf{Synthetic Type (if applicable):} No anomalies typically associated with synthetic or deepfake content were identified using state-of-the-art detection tools.
    \item \textbf{Other Artifacts:} Minor compression artifacts were observed but are consistent with the distribution across social media platforms and do not indicate tampering.
\end{itemize}
\textbf{\# Other Evidence \& Findings}
\begin{itemize}
    \item \textbf{Supporting Sources:}
        \begin{itemize}
            \item Multiple video analysis YAML files (\texttt{ID43-3\_ID43-2}, \texttt{ID43-3\_ID43-3}, \texttt{ID43-3\_ID43-1}) consistently report the explosion at a bridge in Dnipro, Ukraine, with details such as missile impact, smoke plume, and associated timestamps.
            \item The Twitter post (\texttt{https://twitter.com/cesarnews4/status/1521949838916521984}) serves as a primary reference, with additional reshares corroborating the event across platforms.
        \end{itemize}
    \item \textbf{Cross-Checking Information:}
        \begin{itemize}
            \item The location details (Naberezhna Zavodska, Dnipro) are cross-referenced with urban landmark images and satellite data of Dnipro, confirming the recognizable bridge structure.
            \item Timestamps and temporal clues (morning/evening light, timestamp overlay '04/05/2022 19:58:37') align consistently across the different video sources.
        \end{itemize}
    \item \textbf{Other Info:}
        \begin{itemize}
            \item The metadata info regarding a missile strike on a bridge is supported by visual evidence in terms of large explosion flash, resultant fireball, and rising smoke observed in the videos.
            \item The violent nature and high level of damage are further emphasized by the associated violence classification tags and detailed description across multiple evidentiary reports.
        \end{itemize}
\end{itemize}
\end{prompttemplate}
\end{figure*}
To demonstrate the effectiveness of our multimedia verification system, we present a comprehensive case study using sample \texttt{ID43-3} from the \textit{ACMMM25 - Grand Challenge on Multimedia Verification}. 

\begin{figure}[h!]
    \centering
    \includegraphics[width=0.5\linewidth]{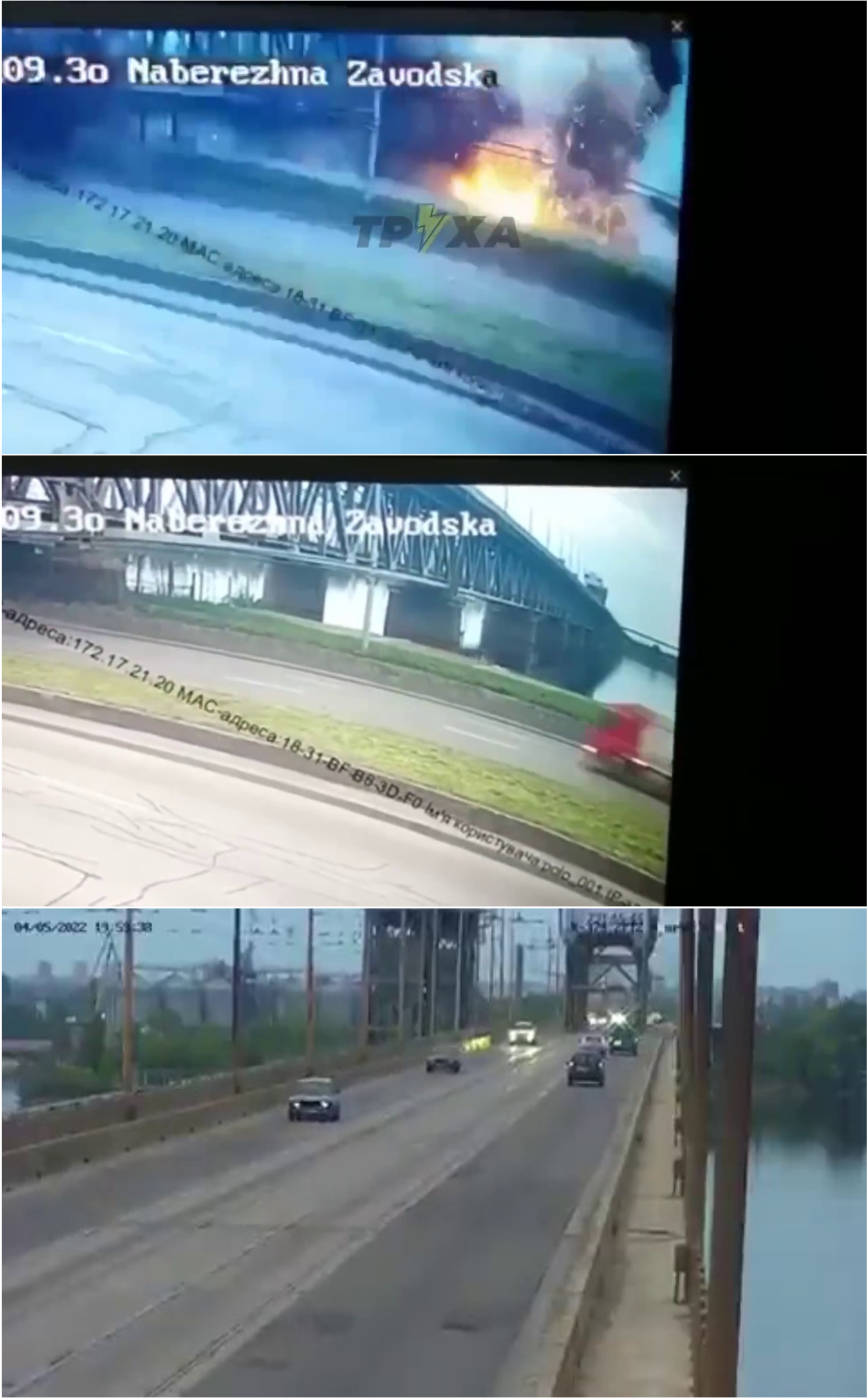}
    \caption{Detected key frames from raw data \texttt{ID43-3}.}
    \label{fig:frames}
\end{figure}

Figure \ref{fig:frames} shows the key frames extracted by our system's raw data processing stage from the submitted video content. The Frame Extractor component successfully identified the most informative visual elements, including the moment of impact, the resulting explosion, and the aftermath with visible smoke and structural damage. These extracted frames provided the visual foundation for subsequent verification stages. Our system processed this case through the complete verification pipeline, generating a comprehensive verification report that demonstrates the practical application of our methodology. The Deep Researcher Agent successfully employed all four specialized verification tools, with the Verified News Tool extracting critical source details across the four key dimensions: spatial context (Where?), temporal context (When?), attribution context (Who?), and motivational context (Why?). The verification process successfully confirmed the authenticity of the submitted content, classifying it as ``Verified'' based on converging evidence from multiple sources. The system's ability to extract and cross-reference precise geolocation coordinates (48.4647° N, 35.0462° E), exact timestamps (04/05/2022, 19:58:37 local time), and detailed source attribution demonstrates the effectiveness of our multi-modal approach.
The system's comprehensive source analysis is particularly noteworthy, tracing the content's origin to a specific Twitter account and documenting its subsequent distribution across multiple platforms. The forensic analysis component detected no signs of synthetic manipulation or deepfake content, while identifying minor compression artifacts consistent with legitimate social media distribution.

This result validates several key aspects of our proposed multi-agent deep research MLLMs multimedia verification system. The iterative nature of the Deep Researcher Agent enabled systematic evidence gathering from diverse sources, while the system's ability to handle complex, real-world content involving geopolitical events demonstrates its robustness and practical applicability to contemporary misinformation challenges.

\section{Conclusion and Future Works}
This paper presents our multi-agent multimedia verification system for the \textit{ACMMM25 - Grand Challenge on Multimedia Verification}. Our approach combines MLLMs with specialized verification tools through a systematic six-stage pipeline that addresses both technical manipulation detection and contextual verification challenges.
The Deep Researcher Agent's four specialized tools, particularly the verified news processor that extracts spatial, temporal, attribution, and motivational context, demonstrate effective integration of diverse verification approaches. Our evaluation on the challenge dataset cases shows the system can successfully handle complex scenarios involving geopolitical content while maintaining detailed source tracking and generating comprehensive verification reports.

Future work will focus on developing contestable AI multimedia verification systems that bridge automated efficiency with human agency across multiple dimensions.
Our research future works encompass three primary directions: (1) \textbf{Scalability and Integration}, including real-time processing capabilities for cross verification responses \cite{phan2025freem2}, expanded multilingual verification across diverse linguistic contexts, and seamless integration with existing fact-checking workflows through standardized APIs and collaborative interfaces; (2) \textbf{Explainable AI (XAI)} methods that enable dynamic multimodal explanation generation \cite{ijcai2024p1025}, uncertainty quantification with confidence assessments \cite{deuschel2024role}, and source tracking for evidence-based decision making \cite{chundru2021leveraging}; or going beyond explaining through (3) \textbf{Human-Centered Contestable AI} with tiered appeal systems, culturally-adaptive explanation interfaces that account for diverse stakeholder needs, and interactive verification workflows that preserve human oversight while maintaining system efficiency \cite{alfrink2023contestable,schmude2025explainability,nguyen2025heart2mind,karusala2024understanding}.

\bibliographystyle{ACM-Reference-Format}
\bibliography{acmart}

\end{document}